\documentclass[floatfix,rmp,twocolumn,twoside]{revtex4}
\usepackage{graphicx}

\bibpunct{[}{]}{,}{n}{}{,}

\usepackage[english]{babel}
\usepackage{verbatim}
\usepackage{varwidth}
\usepackage{graphicx}
\usepackage{alltt}

\makeatletter
\g@addto@macro\@verbatim\small
\makeatother

\begin{document}

\title{Fence --- An Efficient Parser with Ambiguity Support \\ for Model-Driven Language Specification}
\author{Luis~Quesada, Fernando~Berzal, and Francisco J.~Cortijo\\
  Department of Computer Science and Artificial Intelligence, CITIC, University of Granada, \\
  Granada 18071, Spain \\
  \textit{lquesada@decsai.ugr.es, fberzal@decsai.ugr.es, cb@decsai.ugr.es}
  }

\begin{abstract}
Model-based language specification has applications in the implementation of language processors, the design of domain-specific languages, model-driven software development, data integration, text mining, natural language processing, and corpus-based induction of models.
Model-based language specification decouples language design from language processing and, unlike traditional grammar-driven approaches, which constrain language designers to specific kinds of grammars, it needs general parser generators able to deal with ambiguities.
In this paper, we propose Fence, an efficient bottom-up parsing algorithm with lexical and syntactic ambiguity support that enables the use of model-based language specification in practice.
\end{abstract}

\maketitle
\section{Introduction}

Most existing language specification techniques \cite{Aho1972} require the developer to provide a textual specification of the language grammar.

When the use of an explicit model is required, its implementation requires the development of the conversion steps between the model and the grammar, and between the parse tree and the model instance.
Thus, in this case, the implementation of the language processor becomes harder.

Whenever the language specification is modified, the developer has to manually propagate changes throughout the entire language processor pipeline. These updates are time-consuming, tedious, and error-prone.
This hampers the maintainability and evolution of the language \cite{Kats2010}.

Typically, different applications that use the same language are developed.
For example, the compiler, different code generators, and the tools within the IDE, such as the editor or the debugger.
The traditional language processor development procedure enforces the maintenance of several copies of the same language specification in sync.

In contrast, model-based language specification \cite{Kleppe2007} allows the graphical specification of a language.
By following this approach, no conversion steps have to be developed and the model can be modified as needed without having to worry about the language processor, which will be automatically updated accordingly.
Also, as the software code can be combined with the model in a clean fashion, there is no embedding or mixing with the language processor.

Model-based language specification has direct applications in the following fields:

\begin{itemize}
\item The generation of language processors (compilers and interpreters) \cite{Aho2006}.
\item The specification of domain-specific languages (DSLs), which are languages oriented to the domain of a particular problem, its representation, or the representation of a specific technique to solve it \cite{Fowler2010,Hudak1996,Mernik2005}.
\item The development of Model-Driven Software Development (MDSD) tools \cite{Schmidt2006}.
\item Data integration, as part of the preprocessing process in data mining \cite{Tan2006}.
\item Text mining applications \cite{Turmo2006,Crescenzi2004}, in order to extract high quality information from the analysis of huge text data bases.
\item Natural language processing \cite{Jurafsky2009} in restricted lexical and syntactic domains.
\item Corpus-based induction of models \cite{Klein2004}.
\end{itemize}

However, due to the nature of this specification technique and the aforementioned application fields, the specification of separate elements may cause lexical ambiguities to arise.
Lexical ambiguities occur when an input string simultaneously corresponds to several token sequences \cite{Nawrocki1991}.
Tokens within alternative sequences may overlap.

The \emph{Lamb} lexical analyzer \cite{Quesada2011a} captures all the possible sequences of tokens and generates a lexical analysis graph that describes them all.
In these graphs each token is linked to its preceding and following tokens, and there may be several starting tokens.
Each path in this graph describes a possible sequence of tokens that can be found within the input string.

Our proposal, Fence, accepts as input a lexical analysis graph, and performs an efficient ambiguity-supporting syntactic analysis, producing a parse graph that represents all the possible parse trees.
The parsing process discards any sequence of tokens that does not provide a valid syntactic sentence conforming to the production set of the language specification.
Therefore, a context-sensitive lexical analysis is implicitly performed, as the parsing determines which tokens are valid.

The combined use of a \emph{Lamb}-like lexer and Fence allows processing languages with lexical and syntactic ambiguities, which renders model-based language specification techniques usable.

\section{Background}

Formal grammars are used to specify the syntax of a language \cite{Aho2006}.
Context-free grammars are formal grammars in which the productions are of the form $N \rightarrow (\Sigma \cup N)^{*}$ \cite{Chomsky1956}, where  $N$ is a finite set of nonterminal symbols, none of which appear in strings formed from the grammar; and $\Sigma$ is a finite set of terminal symbols, also called tokens, that can appear in strings formed from the grammar, being $\Sigma$ disjoint from $N$.
These grammars generate context-free languages.

A context-free grammar is said to be ambiguous if there exists a string that can be generated in more than one way.
A context-free language is inherently ambiguous if all context-free grammars generating it are ambiguous.

Typically, language processing tools divide the analysis into two separate phases; namely, scanning (or lexical analysis) and parsing (or syntax analysis).

A lexical analyzer, also called lexer or scanner, processes an input string conforming to a language specification and produces the sequence of tokens found within it.

A syntactic analyzer, also called parser, processes an input data structure consisting of tokens and determines its grammatical structure with respect to the given language grammar, usually in the form of parse trees.

\section{Lexical Analysis with Ambiguity Support}

When using a \emph{lex}-generated lexer \cite{Levine1992}, tokens get assigned a priority based on the length of the performed matches and, when there is a tie in the length, on the order of specification.

Given a language specification that describes the tokens listed in Figure \ref{fig:tokens}, the input string ``\&5.2\& /25.20/'' can correspond to the four different lexical analysis alternatives enumerated in Figure \ref{fig:analysis}, depending on whether the sequences of digits separated by points are considered real numbers or integer numbers separated by points.

\begin{figure}[htb*]
\centering
\begin{varwidth}{\linewidth}
\begin{verbatim}
   (-|\+)?[0-9]+              Integer
   (-|\+)?[0-9]+\.[0-9]+      Real
   \.                         Point
   \/                         Slash
   \&                         Ampersand
\end{verbatim}
\end{varwidth}
\caption{Specification of the token types and associated regular expressions of a lexically-ambiguous language.}
\label{fig:tokens}
\end{figure}

\begin{figure}[htb*]
\begin{itemize}
\item \texttt{\small{Ampersand Integer Point Integer Ampersand Slash Integer Point Integer Slash}}
\item \texttt{\small{Ampersand Integer Point Integer Ampersand Slash Real Slash}}
\item \texttt{\small{Ampersand Real Ampersand Slash Integer Point Integer Slash}}
\item \texttt{\small{Ampersand Real Ampersand Slash Real Slash}}
\end{itemize}
\caption{Different possible token sequences in the input string ``\&5.2\& /25.20/'' due to the lexically-ambiguous language specification in Figure \ref{fig:tokens}.}
\label{fig:analysis}
\end{figure}

\begin{figure*}[p]
\centering
\includegraphics[scale=0.388]{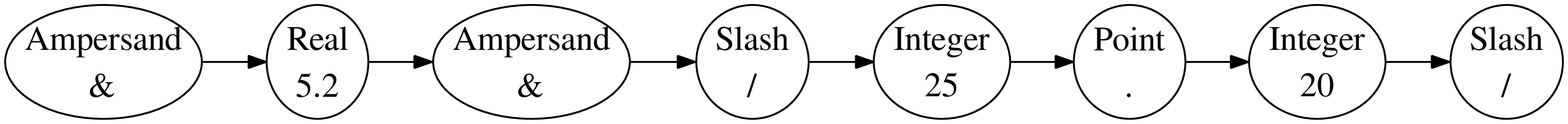}
\caption{Intended lexical analysis.}
\label{fig:e4}
\end{figure*}

The productions shown in Figure \ref{fig:srules1} illustrate a scenario of lexical ambiguity sensitivity.
Depending on the surrounding tokens, which may be either \emph{Ampersand} tokens or \emph{Slash} tokens, the sequences of digits separated by points should be considered either \emph{Real} tokens or \emph{Integer Point Integer} token sequences.
The expected results of analyzing the input string ``\&5.2\& /25.20/'' is shown in Figure \ref{fig:e4}.

\begin{figure}[htb*]
\centering
\begin{varwidth}{\linewidth}
\begin{verbatim}
E ::= A B
A ::= Ampersand Real Ampersand
B ::= Slash Integer Point Integer Slash
\end{verbatim}
\end{varwidth}
\caption{Context-sensitive productions that solve the lexical ambiguities in Figure \ref{fig:analysis}.}
\label{fig:srules1}
\end{figure}

\begin{figure*}[p]
\centering
\noindent \includegraphics[scale=0.388]{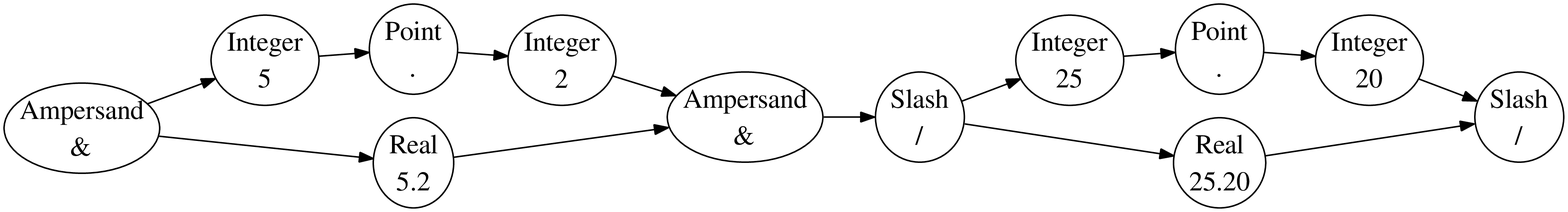}
\caption{Lexical analysis graph, as produced by the \emph{Lamb} lexer.}
\label{fig:e1}
\end{figure*}

\begin{figure*}[p]
\centering
\includegraphics[scale=0.388]{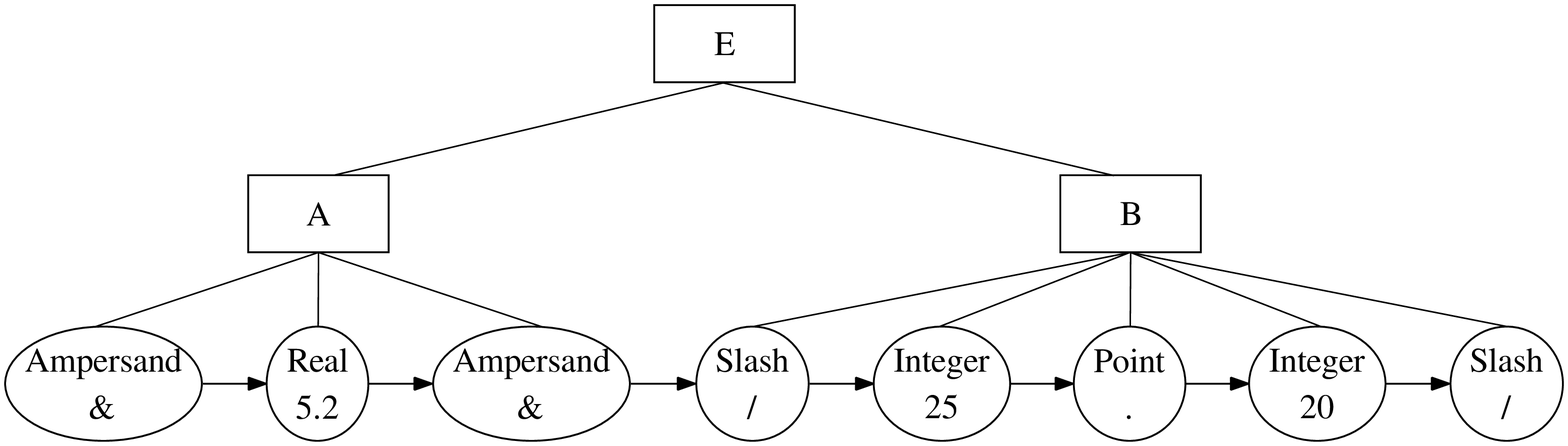}
\caption{Syntactic analysis graph, as produced by applying a parser that supports lexical ambiguities to the lexical analysis graph shown in Figure \ref{fig:e1}.
Squares represent nonterminal symbols found during the parse process.}
\label{fig:e5}
\end{figure*}

The \emph{Lamb} lexer \cite{Quesada2011a} performs a lexical analysis that efficiently captures all the possible sequences of tokens and generates a lexical analysis graph that describes them all, as shown in Figure \ref{fig:e1}.
The further application of a parser that supports lexical ambiguities would produce the only possible valid sentence, which, in turn, would be based on the only valid lexical analysis possible.
The intended results are shown in Figure \ref{fig:e5}.

\begin{figure*}[p]
\centering
\includegraphics[scale=0.388]{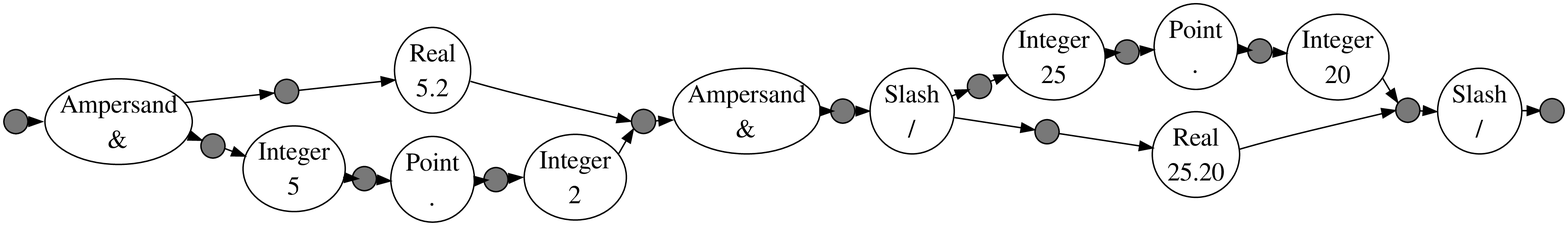}
\caption{Extended lexical analysis graph corresponding to the lexical analysis graph shown in Figure \ref{fig:e1}. Gray nodes represent cores}
\label{fig:e6}
\end{figure*}

\begin{figure*}[p]
\centering
\includegraphics[scale=0.388]{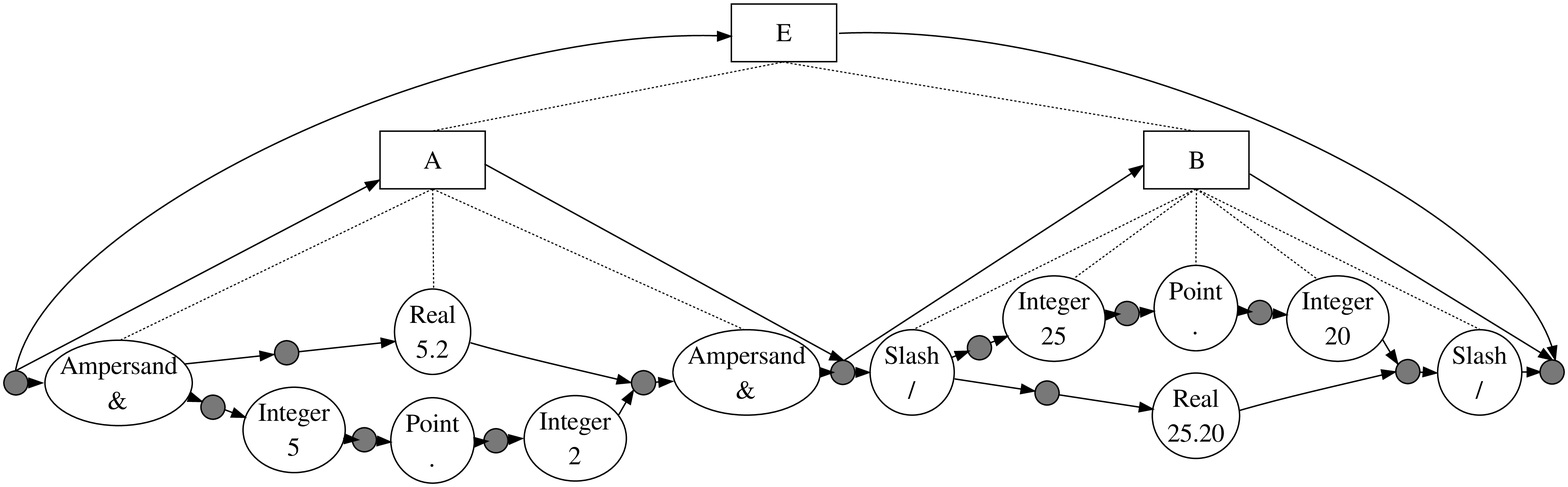}
\caption{Extended syntax analysis graph corresponding to the extended lexical analysis graph shown in Figure \ref{fig:e6}. 
Squares represent nonterminal symbols found during the parse process.}
\label{fig:e7}
\end{figure*}

\section{Syntactic Analysis with Ambiguity Support}

Traditional efficient parsers for restricted context-free grammars, as the LL \cite{Rosenkrantz1970}, SLL, LR \cite{Knuth1965}, SLR, LR(1), or LALR parsers \cite{Aho2006}, do not consider ambiguities in syntactic analysis, so they cannot be used to perform parsing in those cases. The efficiency of these parsers is $O(n)$, being $n$ the token sequence length.

Existing parsers for unrestricted context-free grammar parsing, as the CYK parser \cite{Younger1967,Kasami1969} and the Earley parser \cite{Earley1983}, can consider syntactic ambiguities. The efficiency of these parsers is $O(n^3)$, being $n$ the token sequence length.

In contrast to the aforementioned techniques, our proposed parser, Fence, is able to efficiently process lexical analysis graphs and, therefore, consider lexical ambiguities.
It also takes into consideration syntactic ambiguities.

Fence produces a parse graph that contains as many starting initial grammar symbols as different parse trees exist.

\subsection{Extended Lexical Analysis Graph}

In order to efficiently perform the parsing, Fence uses an extended lexical analysis graph that stores information about partially applied rules, namely handles, in data structures, namely cores.

Given a sequence of symbols $T = t_1 ... t_n$ as the right hand side of a production rule, a dotted rule is a pair $(production,pos)$, where $0\leq pos\leq n$.

A handle is a dotted rule associated to a starting position in the input string.

A core is a set of handles.

In an extended lexical analysis graph, tokens are not linked to their preceding and following tokens, but to their preceding and following cores.
Cores are, in turn, linked to their preceding and following token sets.
For example, the extended lexical analysis graph corresponding to the lexical analysis graph in Figure \ref{fig:e1} is shown in Figure \ref{fig:e6}.

As cores represent a starting position in the input string, handles are a dotted rule associated to a starting core.

Each handle could be used to make the analysis progress (namely, \emph{SHIFT} actions in LR-like parsers) or perform a reduction (namely, \emph{REDUCE} actions in LR-like parsers).

A shift action needs to be performed associated to a source core and a target core.
Applying the shift action to a handle involves creating a new handle in each target core that follows the symbols that follow the source core.

A reduction action needs to be performed associated to a $start$ core and an $end$ core.

\subsection{Parsing Algorithm}

The algorithm uses a global matched handle pool, namely \emph{hPool}, that contains handles associated to the next symbol they can match.

The first step of our algorithm converts the input lexical analysis graph into an extended lexical analysis graph.

This conversion is performed by completing the graph with a \emph{first} core, which links to the tokens with an empty preceding token set; a \emph{last} core, which is linked from the tokens with an empty following token set; and, for each one of the other tokens, a core that precedes it. 
Links between tokens are then converted to links from tokens to the cores preceding each token of their following token set and vice versa.

The second step of our algorithm performs the parsing, by progressively applying productions and storing handles in cores.

First, the productions with an empty right hand side are removed from the grammar and their left hand side element is stored in a set named \emph{epsilonSymbols}.

The \emph{addProd} procedure described in Figure \ref{fig:codeaddrule} generates a handle conforming to a production and a starting right hand side element index, adds it to a core and, for each symbol in the following symbol set of that core that matches the current production element, adds a handle to the production pool with an anchor to that symbol. 
It also considers productions with an empty right hand side: if an element is in the \emph{epsilonSymbols} set, both the possibilities of it being reduced or not by that production are considered, that is, if an element corresponds is in the \emph{epsilonSymbols} set, a new handle that skips that element is added to the same core. 
It should be noted that this process is iterative, as many sucessive elements of the right hand side of a production could be in the \emph{epsilonSymbols} set.

\begin{figure}[htb]
\begin{verbatim} 
procedure addProd(Prod p,int index,Core c,
                  Core start,Symbol[] contents):
  do:
    h = new Handle(p,index,start)
    c.handles.add(h)
    if index < p.right.size:
      for each Symbol s in c.following:
        if s.type == p.right[index].type:
          hPool.add({new Handle(p,index,start,
                     contents+s)})
    index++
    contents.add(null) // epsilon symbol case
  while index < p.right.size &&
          epsilonSymbols.has(p.right[index-1].type)
\end{verbatim}
\caption{The \emph{addProd} procedure pseudocode.}
\label{fig:codeaddrule}
\end{figure}

\begin{figure}[htb]
\begin{verbatim}
for each Prod p in prodSet:
  for each Core c in coreSet:
    flag = false
    for each Token t in c.following:
      if t is in p.selectSet:
        flag = true
    if flag == true:
      addProd(p,0,c,c,null)
\end{verbatim}
\caption{Core initialization.}
\label{fig:codeinit}
\end{figure}

\begin{figure}[htb]
\begin{verbatim}
  while hPool is not empty:
    {h,symbol} = hPool.extract()
    if h.index == h.prod.right.size-1:
      // Production matched all its elements.
      // i.e. Reduction
      s = new Symbol(h.prod.left.type,h.contents)
      h.startCore.add(s)
      s.preceding.add(h.start)
      for each Core c in h.following:
        c.preceding.add(s)
        s.following.add(c)
      for each Handle h in h.startCore that
                             is waiting for s.type:
        hPool.add({new Handle(h.prod,h.index,
                   h.start,contents),s})
    else: // i.e. Shift
      for each Core c in h.following:
        addProd(h.prod,h.index+1,c,h.start)
\end{verbatim}
\caption{Pseudocode of the parsing algorithm.}
\label{fig:codeprocess}
\end{figure}

The SELECT set contains all of the terminal symbols first produced by the production.

The parser is initialized by generating every possible handle that would match the first right hand side element of a rule, and adding it to every core whose following tokens are in the SELECT set of the production, as shown in Figure \ref{fig:codeinit}. 

The parsing process consists on iteratively extracting handles from \emph{hPool} and matching them with the following, already known, symbol. 
The handles derived from that match are added to the corresponding cores and, for each symbol in the following set of symbols of the core that matches the next unmatched element of the production, to the rule pool.

In case all the elements of a production match a sequence of symbols, a new symbol is generated by reducing them, and added to the rule start core. 
If a new added symbol only has the \emph{first} core in its preceding core set and the \emph{last} core in its following core set, and it is an instance of the initial symbol of the grammar, it is added to the parse graph starting symbol set. 
The pseudocode for this process is shown in Figure \ref{fig:codeprocess}.

It should be noted that handles are never removed from the cores when shift actions are performed.
This allows generating parse trees that consist of nonterminal symbols found later in the parsing process.

The result is an extended parse graph, as the one shown in Figure \ref{fig:e7}.

In the last step of the algorithm, all the cores are stripped off the graph and the symbols are linked back to their new preceding and following symbol sets, in order to produce the output syntax analysis graph.

\subsection{Efficiency Analysis}

The following efficiency analysis does not consider enumerating all the different parse trees, which the pseudocode shown in section 4.2 does and has an exponential order of efficiency.
Instead, it considers a simplified theoretical parsing process.

Let $n$ denote the input string length, $p$ the number of productions of the grammar, $l$ the maximum length of a production (the number of symbols in its right hand side), and $s$ the number of terminal symbols of the grammar.

We define $d$ as the dimension of a grammar, that is, the sum of the number of symbols that appear in the right hand side of the productions of the grammar.

Nonterminal symbols, which are created whenever a reduction is performed, can be defined as tuples $(X,start,end)$, being $start$ the start core identifier and $end$ the end core identifier, where $end>=start$.
A nonterminal symbol corresponds to a single parse tree if the grammar has no ambiguities, and may correspond to a set of parse trees if the grammar has lexical or syntactic ambiguities.

If the input string is successfully parsed, the result will be $(S,1,n)$, being $S$ the initial symbol of the grammar.

An extended lexical analysis graph contains a number of tokens that is conditioned by the input length and the presence of lexical ambiguities.
It also contains a number of cores that is conditioned by the number of tokens.

Each core will store a number of handles that is conditioned by the grammar power of expression and the presence of lexical ambiguities.

\subsubsection{Parsing LR Grammars without Lexical Ambiguities}

An input string length of $n$ means a maximum of $n$ tokens can be found, in the absence of lexical ambiguities.
A lexical analysis graph with $n$ tokens will contain a maximum of $n$ cores.

In this case, each core can initially store up to $l$ handles, as symbols that appear in the left hand side of productions with an empty right hand side may be skipped during the initialization, and all the different handles that represent these possibilities have to be considered.
Thus, $n\cdot l$ handles may initially exist.

Each handle can cause, at most, $l$ shift actions, each of which would generate, at most, a single new handle.
Each shift action can be performed in constant time.

Therefore, a maximum of $n\cdot l\cdot (1+l)$ handles can be generated.
Each handle can be generated in constant time.

Also, each handle can cause, at most, a reduction, which would generate a single nonterminal symbol.
This reduction can be performed in constant time.

Thus, the order of efficiency of our algorithm in this case is $O(n\cdot l^2)$.

\subsubsection{Parsing LR Grammars with Lexical Ambiguities}

An input string length of $n$ means a maximum of $n\cdot s$ tokens can be found, in the presence of lexical ambiguities.
A lexical analysis graph with $n\cdot s$ tokens will contain a maximum of $n\cdot s$ cores.

In this case, each core can initially store up to $l$ handles, as symbols that appear in the left hand side of productions with an empty right hand side may be skipped during the initialization, and all the different handles that represent these possibilities have to be considered.
Thus, $n\cdot s\cdot l$ handles may initially exist.

Each handle can cause, at most, $l$ shift actions, each of which would generate up to $s$ handles.
This sums up to $s\cdot l$ handles.

Therefore, a maximum of $n\cdot s\cdot l\cdot (1+s\cdot l)$ handles can be generated.
Each handle can be generated in constant time.

Also, each handle can cause, at most, a reduction, which would generate a single nonterminal symbol.
This reduction can be performed in constant time.

Thus, the order of efficiency of our algorithm in this case is $O(n\cdot s^2\cdot l^2)$.
The memory it uses has an order of $O(n\cdot s^2\cdot l^2)$, too.

Considering $s$ as a constant, the order of efficiency of our algorithm is $O(n\cdot l^2)$.
The reason $s$ appears in the order of efficiency is that lexical ambiguities, which could be solved by using a parser with syntactic ambiguity support and rewriting the grammars in order to model them as syntactic ambiguities, are considered during a previous lexical analysis, thus generating tokens which, otherwise, would be nonterminal symbols.

\subsubsection{Parsing CFG Grammars without Lexical Ambiguities}

An input string length of $n$ means a maximum of $n$ tokens can be found, in the absence of lexical ambiguities.
A lexical analysis graph with $n$ tokens will contain a maximum of $n$ cores.

In this case, each core can initially store up to $d$ handles, as symbols that appear in the left hand side of productions with an empty right hand side may be skipped during the initialization, and all the different handles that represent these possibilities have to be considered.
Thus, $n\cdot d$ handles may initially exist.

Each handle can cause, at most, $l$ shift actions, each of which would generate, at most, a single new handle.
Each shift action can be performed in constant time.

Therefore, a maximum of $n\cdot d\cdot (1+l)$ handles can be generated.
Each handle can be generated in constant time.

Also, each handle can cause, at most, a reduction, which would generate a single nonterminal symbol.
This reduction can be performed in constant time.

Thus, the order of efficiency of our algorithm in this case is $O(n\cdot d\cdot l)$.
The memory it uses has an order of $O(n\cdot d\cdot l)$, too.

\subsubsection{Parsing CFG Grammars with Lexical Ambiguities}

An input string length of $n$ means a maximum of $n\cdot s$ tokens can be found, in the presence of lexical ambiguities.
A lexical analysis graph with $n\cdot s$ tokens will contain a maximum of $n\cdot s$ cores.

In this case, each core can initially store up to $d$ handles, as symbols that appear in the left hand side of productions with an empty right hand side may be skipped during the initialization, and all the different handles that represent these possibilities have to be considered.
Thus, $n\cdot s\cdot d$ handles may initially exist.

Each handle can cause, at most, $l$ shift actions, each of which would generate up to $s$ handles.
This sums up to $s\cdot l$ handles.

Therefore, a maximum of $n\cdot s\cdot d\cdot (1+s\cdot l)$ handles can be generated.
Each handle can be generated in constant time.

Also, each handle can cause, at most, a reduction, which would generate a single nonterminal symbol.
This reduction can be performed in constant time.

Thus, the order of efficiency of our algorithm in this case is $O(n\cdot s^2\cdot d\cdot l)$.
The memory it uses has an order of $O(n\cdot s^2\cdot d\cdot l)$, too.

Considering $s$ as a constant, the order of efficiency of our algorithm is $O(n\cdot d\cdot l)$.
The reason $s$ appears in the order of efficiency is that lexical ambiguities, which could be solved by using a parser with syntactic ambiguity support and rewriting the grammars in order to model them as syntactic ambiguities, are considered during a previous lexical analysis, thus generating tokens which, otherwise, would be nonterminal symbols.

\section{Conclusions and Future Work}

Model-based language specification decouples language design from language processing.
Languages specified using such technique may be lexically and syn-tactically-ambiguous.
Thus, general parser generators able to deal with ambiguities are needed.

We have presented Fence, an efficient bottom-up parsing algorithm with lexical and syntactic ambiguity support that enables the use of model-based language specification in practice.

Fence accepts a lexical analysis graph as input, performs a syntactic analysis conforming to a grammar specification, and produces as output a compact representation of a set of parse trees.

We plan to apply model-based language specification in the implementation of language processor generators, model-driven software development, data integration, corpus-based induction of models, text mining, and natural language processing.

\bibliographystyle{plain}
\bibliography{doc}

\end{document}